\pdfoutput=1

\documentclass[11pt]{article}

\usepackage[]{acl}

\usepackage{times}
\usepackage{latexsym}
\usepackage{graphicx}
\usepackage{url}
\usepackage[T1]{fontenc}

\usepackage[utf8]{inputenc}

\usepackage{microtype}

\usepackage{algorithm}
\usepackage{algorithmic}
\usepackage{multirow, tabularx}
\usepackage{array}
\usepackage{booktabs}
\usepackage{setspace}
\usepackage{xcolor}
\usepackage{makecell}

\title{MICO: A Multi-alternative Contrastive Learning Framework for \\ Commonsense Knowledge Representation}

\author{{\bf Ying Su}$^1$, {\bf Zihao Wang}$^1$, {\bf Tianqing Fang}$^1$, {\bf Hongming Zhang}$^{2}$, \\ {\bf Yangqiu Song}$^1$,
{\bf Tong Zhang}$^1$ \\
$^{1}$HKUST, $^{2}$Tencent AI lab, Seattle \\
 \texttt{\{ysuay,zwanggc\}@connect.ust.hk}, 
 \texttt{\{tfangaa,yqsong\}@cse.ust.hk}, \\
 \texttt{hongmzhang@global.tencent.com},
 \texttt{tongzhang@ust.hk}
\\
}

\begin{document}
\maketitle

\begin{abstract}


Commonsense reasoning tasks such as commonsense knowledge graph completion and commonsense question answering require powerful representation learning.
In this paper, we propose to learn commonsense knowledge representation by MICO, a \textbf{M}ulti-alternative contrast\textbf{I}ve learning framework on \textbf{CO}mmonsense knowledge graphs (MICO).
MICO generates the commonsense knowledge representation by contextual interaction between entity nodes and relations with multi-alternative contrastive learning. 
In MICO, the head and tail entities in an $(h,r,t)$ knowledge triple are converted to two relation-aware sequence pairs (a premise and an alternative) in the form of natural language.
Semantic representations generated by MICO can benefit the following two tasks by simply comparing the distance score between the representations: 1) zero-shot commonsense question answering task; 2) inductive commonsense knowledge graph completion task. Extensive experiments show the effectiveness of our method.
\end{abstract}


\section{Introduction}

Commonsense reasoning is a fundamental problem in artificial intelligence. Recently in the NLP field, much attention has been paid to commonsense reasoning in the following two aspects. 
First, more commonsense knowledge graphs (CKGs) \cite{sap2019atomic, fang2021discos} were developed to support new types of reasoning tasks, such as commonsense knowledge graph completion (CKGC)~\cite{malaviya2020commonsense}. 
Another way to evaluate machine learning models' commonsense reasoning capabilities is using commonsense question answering (CQA) tasks~\cite{zellers2018swag, sap2019social, bisk2020piqa}.
Existing approaches to deal with the above problems commonly involve fine-tuning large pre-trained language models, such as BERT \cite{kenton2019bert}, RoBERTa \cite{liu2019roberta}, and GPT2 \cite{radford2019language}, by either incorporating the entire knowledge base for CKGC ~\cite{yao2019kg,bosselut2019comet} or injecting the knowledge base to provide background knowledge for zero-shot CQA \cite{banerjee2020self, bosselut2021dynamic, ma2021knowledge}. 


In fact, both CKGC and zero-shot CQA can be formulated in a unified way, where a question can be constructed based on the {\it head entity} and {\it relation} in a knowledge graph, and then finding the {\it tail entity}, which is regarded as an answer, based on the constructed question.
In this way, incorporating the entire knowledge base for CKGC and injecting the KG in pre-trained LMs for zero-shot CQA can be unified as a semantic matching problem, where a powerful representation learning for the matching becomes the most important problem. 
This also means that, after we unify them for CKGC and CQA in the same way, we can perform zero-shot CQA by simply leveraging the model finetuned on the entire CKGs for CKGC.


Existing commonsense-related representation learning usually leverage a CKG embedding framework \cite{malaviya2020commonsense, wang2021inductive}, or fine-tuning a generative language model~\cite{bosselut2019comet}. However, they were not aware of the challenges that a typical CKG brings. First, in a typical CKG, such as ConceptNet~\cite{liu2004conceptnet} and ATOMIC~\cite{sap2019atomic}, nodes are loosely structured free-from texts, which means that previous embedding based on negative sampling cannot substantially support sufficient training because of sparsity. On the other hand, a generative model can only take positive examples for training so the capability of determining the negative answers is limited.

\begin{figure}[t]
\centering
\includegraphics[scale=0.55, trim={0.5cm 0 0 0}]{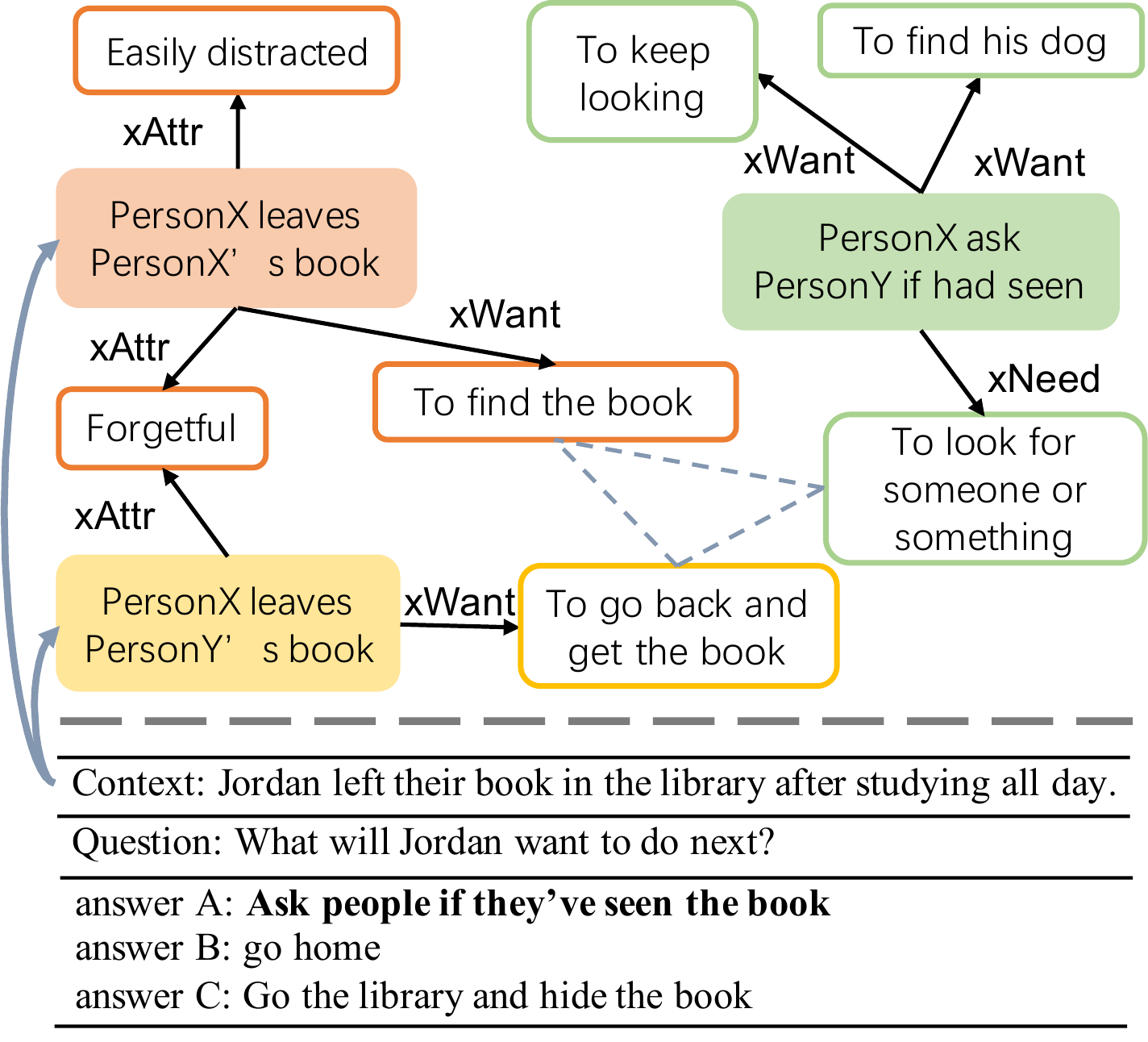}
\caption{An example from a CQA task (SIQA \cite{sap2019social}) and related knowledge in a CKG (ATOMIC).}
\vskip-1.5em
\label{fig:exp}
\end{figure}

In this paper, we propose a new framework called MICO, a \textbf{M}ulti-alternative contrast\textbf{I}ve learning framework for \textbf{CO}mmonsense knowledge representation. The representations can benefit across tasks by easily calculating the semantic distances with unified vector representations. In this way, though many distinct nodes may have similar concepts \cite{wang2021inductive}, they are still close in the semantic space. Figure \ref{fig:exp} shows an example of this advantage on CQA.  The entity node \textit{PersonX ask if PersonY had seen} related to the right answer is not directly connected to the nodes \textit{PersonX leaves PersonX's book} or \textit{PersonX leaves PersonY's book} but these nodes share similar tail entity nodes. Therefore, the right answer can be found by semantic matching as it is close to the given context and question in the semantic space.

To unify the form of CKGC and CQA, we follow the idea in COPA \cite{roemmele2011choice} where commonsense causal reasoning can be evaluated by selecting the most plausible alternatives given the premise. We first converts the knowledge triplets $(h,r,t)$ into sequence pairs $(P, A)$ ($P$ for premise and $A$ for alternative). MICO then encodes the sequence pairs into embeddings and measures their distance by a similarity function as we assume the representations of related knowledge lie close in the embedding space. Furthermore, we enhance the representation learning by a contrastive loss with sufficient sampling over the sparse CKG. The alternative from the same triplet is a positive sample to the premise under the contrastive learning framework. Alternatives from other knowledge triples with different premises are negative samples. MICO also takes consideration of the structure from CKGs, where one head node $h$ may connect to several tail nodes $t$ under the same relation $r$. MICO dynamically selects a hard alternative from multi-alternatives for a premise during training.

Experiments on two typical commonsense knowledge graphs and two types of tasks, zero-shot CQA and inductive CKGC, demonstrate the effectiveness of our methodology. Our code is open-resourced.\footnote{\url{https://github.com/HKUST-KnowComp/MICO}}

\section{Related Work}

\subsection{Commonsense Question Answering}

Background knowledge is necessary for commonsense question answering tasks. Many researches resort to knowledge bases for background knowledge. The works towards this direction can be mainly classified into two streams: incorporating the knowledge base for zero-shot CQA \cite{yang2019enhancing, banerjee2020self, bosselut2021dynamic, ma2021knowledge} or retrieving the related knowledge from the knowledge base for task-specific CQA \cite{paul2019ranking, lin2019kagnet, feng2020scalable, lv2020graph, yasunaga2021qa, xu2021fusing, zhang2021greaselm}.


Among the works in incorporating the knowledge base for zero-shot CQA, COMET-DynaGen \cite{bosselut2021dynamic} aggregates all paths of generated commonsense knowledge to the answers from commonsense transformer COMET \cite{bosselut2019comet} trained on CKGs. KTL \cite{banerjee2020self} encodes the knowledge triplets from CKGs into pre-trained LMs by learning triplet representation, aiming to complete a knowledge triplet given the other two. Unlike KTL, we target enhancing the relation-aware representation learning in the form of natural language sequence pairs. 

\begin{figure*}[t]
\centering
\includegraphics[scale=0.5]{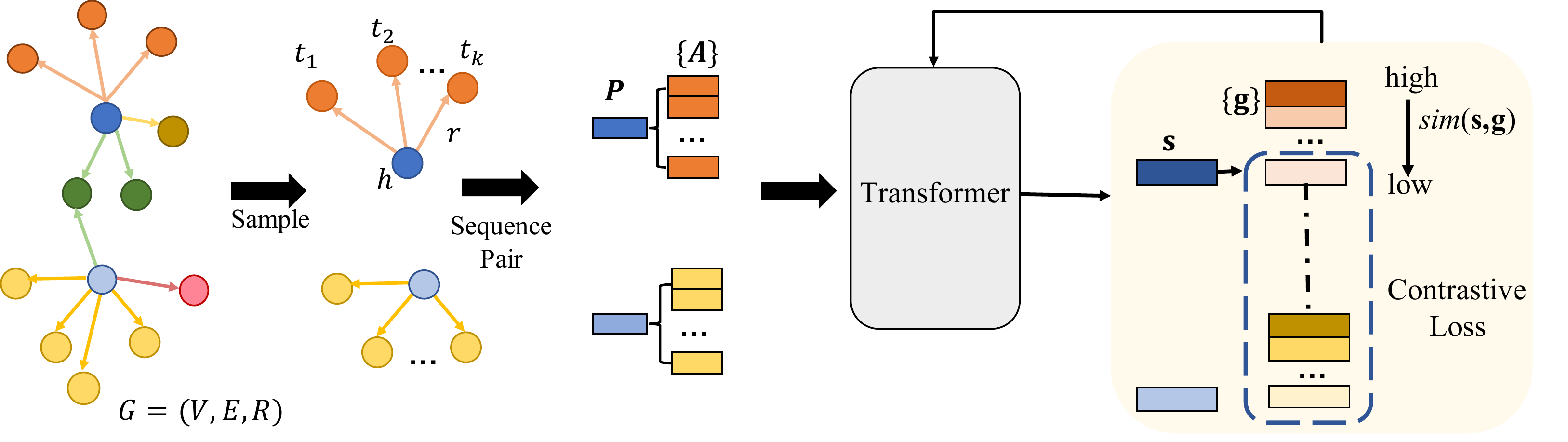}
\caption{Overview framework of MICO. The knowledge triplets are first sampled and converted into sequence pairs. A transformer block encodes the sequence pairs into knowledge embeddings. The contrastive loss updates the parameters of the transformer based on the contrast between the selected positive and negative tail sequences given a head sequence.}
\label{fig:model}
\end{figure*}

\subsection{CKG Knowledge Representation}

Knowledge representation from knowledge graphs (KGs) has significantly progressed and benefited the KG completion task. Typical methods for KG completion tasks are mainly embedding-based, which utilize the structural information observed in the knowledge triplets \cite{nickel2011three, bordes2013translating, wang2014knowledge, trouillon2016complex, toutanova2015representing, sun2019rotate}. Recent researches also show that external information, such as the textual description of nodes or relation descriptions, can help boost the performance on the task, like ConvE \cite{dettmers2018convolutional} and ConvTransE \cite{shang2019end}. To transfer the knowledge from pre-trained LMs into knowledge graph completion, KG-BERT \cite{yao2019kg} further utilize the pre-trained LMs to learn context-aware embeddings.

However, unlike previous KGs \cite{miller1995wordnet, bollacker2008freebase}, commonsense knowledge graphs, e.g., ConceptNet and ATOMIC, have unique challenges towards the completion task. The nodes in CKGs are non-canonicalized and free-from text, resulting in magnitude larger and sparser graphs \cite{malaviya2020commonsense}. To address this problem, previous works extract entity and relation representation by pre-trained LMs and graph structure representation by graph neural networks GCN \cite{kipf2017semi} to enhance the generalizability over entity nodes \cite{malaviya2020commonsense, wang2021inductive}.
Instead of fusing representation from local subgraph structures in this paper, we focus on utilizing the contrast information between knowledge triplet contexts.

\begin{figure}[t]
\centering
\includegraphics[scale=0.55, trim={0.5cm 0 0 0}]{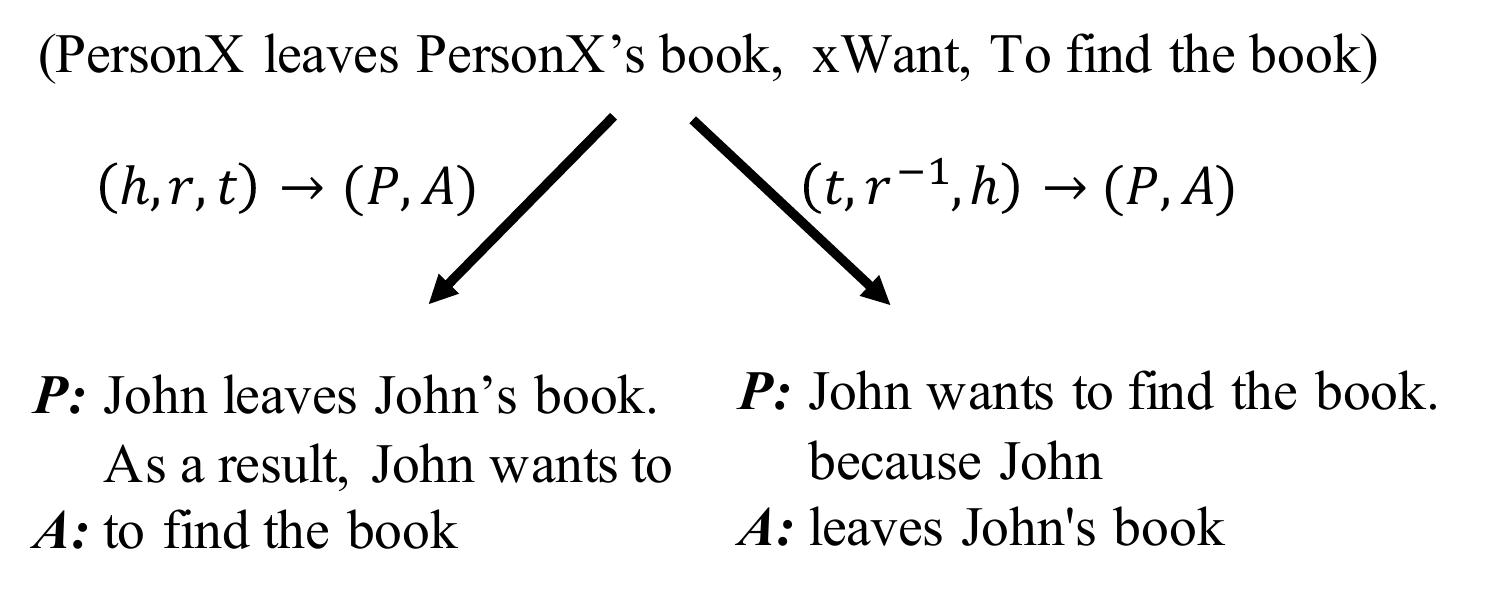}
\caption{An example of converting knowledge triplet $(h,r,t)$  to sequence pairs $(P, A)$ from ATOMIC. $r^{-1}$ is the reverse relation of $r$. }
\label{method:1}
\end{figure}

\section{Methodology}

In this section, we introduce the terminologies and algorithms, and show the framework in Figure \ref{fig:model}.

\subsection{Knowledge Triplets to Sequence Pairs}

A commonsense knowledge graph is denoted as $G=\{V, E, R\}$, where $V$ is the set of entities, $E$ is the set of edges, and $R$ is the set of relations. Knowledge triplet $e \in E$ is composed by $(h, r, t)$ where head entity $h$ and tail entity $t$ are entities from $V$ and $r$ is from $R$. $h$ and $t$ are connected by $r$. Each entity comes with a free-form text description. 

To convert the knowledge triplet into sequence pairs as inputs to MICO, we substitute the relation with human-readable language templates and connect it to entities. Typically relations are represented as specific words or short phrases in the CKG, for example, \textit{xWant} from ATOMIC and \textit{AtLocation} from ConceptNet. Following \citet{hwang2021comet}, we design natural language templates to replace the original relations and connect them to entities, forming context-aware sequences.
An example from ATOMIC is shown in Figure \ref{method:1}. The template for \textit{xWant} would be \textit{as a result, PersonX wants}. We also design a template for its reverse version $r^{-1}$ so it can be connected to the tail entity to form an additional sequence pair. Details of substitute templates  are listed in Appendix~\ref{sec:appendix1}. We denote the newly constructed sequence pair as $(P, A)$. For a premise $P$, there may be multiple alternatives connected to it, denoted as $\{A_{1}, A_{2}, A_{3}, ...\}$.  


\subsection{MICO}

MICO is a multi-alternative contrastive learning framework for commonsense knowledge representation with knowledge sequence pairs as inputs. Recent researches have greatly progressed in sequence representation learning by contrastive learning \cite{carlsson2020semantic, he2020momentum, gao2021simcse}, in which positive sequence pairs are considered semantic related and are close neighbors in the embedding space. MICO follows the idea and minimizes the distance between the premise $P$ and its connected alternative ${A}$.

First, a transformer encodes the constructed sequence pairs $(P, A)$ into embeddings to extract initial representations. Specifically, when using BERT as the transformer, BERT-specific start and end tokens are padded to the input sequence: $x=x_0,x_1,...,x_n$ is converted to $x=[CLS],x_0,x_1,...x_n, [SEP]$. The sequence pairs are then transformed to token pairs as $P_{\rm tok}$ and $A_{\rm tok}$ by a transformer tokenizer. To get the initial representation, a transformer encoder encodes the token pairs as:
\begin{equation}
    E_p = \textnormal{Encoder}(P_{\rm tok}), \\
    E_a = \textnormal{Encoder}(A_{\rm tok}),
\end{equation}
where $E_h$ and $E_t$ are hidden states of the last layer.
The representation of the hidden state for the [CLS] token is used as the representation of the input sequence. For the head and tail sequences, representations are:

\begin{equation}
    \textbf{s} = E_p[0], \\
    \textbf{g} = E_a[0].
\end{equation}

As we assume the sequence pairs lie close in the embedding space, we use a similarity function to measure the distance between sequence pairs. The function $f$ can be cosine similarity or dot product:
\begin{equation}
    sim(\mathbf{s}, \mathbf{g}) = f(\mathbf{s}, \mathbf{g}).
\end{equation}

For a premise $P$, its paired alternative $A$ is a positive sample while alternatives paired with other premises are negative samples in the same batch during training. To minimize the semantic distance between the $i$-th sequence pair in a batch, the contrastive loss is:

\begin{equation}
  \ell_{i}=-\log \frac{e^{sim(\mathbf{s_i}, \mathbf{g^{+}_{i}}) / \tau}}{\sum_{j=1}^{N} e^{sim(\mathbf{s_{i}}, \mathbf{g^{+}_{j}}) / \tau}, }  
\end{equation}
where $N$ is the batch size and $\tau$ is the temperature parameter.

Many research efforts aim to improve representation learning by generating multiple views for the same sample as data augmentation in multi-view contrastive learning approaches \cite{bachman2019learning, tian2020contrastive, niu2022cake}. In CKG, a sequence head $P$ may have multiple positive tails $\{A_{1}, A_{2}, A_{3}, ...\}$. Inspired by multi-view contrastive learning, we propose a multi-alternative framework to utilize the multiple positive alternatives for improving the learning of commonsense knowledge representation. Specifically, we dynamically sample a hard positive alternative from multiple alternatives during training.

The representations generated for the premise with its alternatives are $\mathbf{s}$ and $\{\mathbf{g^{+}_{1}}, \mathbf{g^{+}_{2}}, \mathbf{g^{+}_{3}}, ...\}$. Among the multiple positive alternatives, the one with the largest distance to the premise is selected as the hard positive. Because we aim to minimize the semantic similarity between the premise and alternatives, selecting the alternative with the least similarity would increase the training loss: 
\begin{equation}
    \mathbf{g_{p}} = \min \{sim(\mathbf{s}, \mathbf{g^{+}_{1}}), ..., sim(\mathbf{s}, \mathbf{g^{+}_{k}}) \},
\end{equation}
where $k$ is the number of candidate alternatives during training. The new contrastive loss for $i$-th sequence pairs during training is:
\begin{equation}
\small
L_{i}=-\log \frac{e^{sim(\mathbf{s_i}, \mathbf{g_p}) / \tau}}{ e^{sim(\mathbf{s_i}, \mathbf{g_p}) / \tau} + \delta_{ij}\sum\limits_{j=1}^{N}{\sum\limits_{o=1}^{k}e^{sim(\mathbf{s_i}, \mathbf{g^{+}_{j,o}}) / \tau}}},
\end{equation}
where $\delta_{ij} \in\{0,1\}$ is an indicator that equals 1 if $i \neq j$, and $\mathbf{{g}^{+}_{j,o}}$ is the $o$-th positive tail of $j$-th sample in the batch.

\section{Experiments}

In this section, we first introduce the CKGs used as knowledge sources, and then two kinds of evaluation tasks (zero-shot CQA and inductive CKGC). Finally, we introduce baseline methods for the two tasks separately.

\begin{table*}[t]
    \renewcommand{\arraystretch}{1.2}
    \small
    \centering
    \begin{tabular}{lccccccc}
    \toprule
    Dataset & Entities & Relations & Train Pair & Valid Pair & Test Pair & Avg Degree & Avg Words   \\
    \midrule
    ConceptNet & 78,334 & 34 & 163,840 & 19,590 & 19,592 & 1.87 & 3.93 \\
    ATOMIC & 304,388 & 9 & 1,221,072 & 48,710 & 48,972 & 2.52 & 6.12 \\
    \bottomrule
    \end{tabular}
    \caption{Distribution of train, valid, and test sequence pairs from ConceptNet and ATOMIC. \textit{Avg Degree} is the average number of tail sequence connected to head sequence and \textit{Avg Words} is the average words number for head sequence and tail sequence.}
    \label{cskb:1}
    \vskip-1.0em
\end{table*}

\subsection{CKG}
We conduct experiments on two typical commonsense knowledge graphs, ConceptNet \cite{speer2017conceptnet} and ATOMIC \cite{sap2019atomic}. \\

\noindent \textbf{ConceptNet}. ConceptNet has been the most fundamental commonsense knowledge graph over the past decade \cite{liu2004conceptnet}. CN-100K was built on the knowledge triplets in ConceptNet and first introduced in \citet{li2016commonsense}. It contains Open Mind Common Sense (OMCS) \cite{singh2002public} entries in the ConceptNet5 \cite{speer2017conceptnet}. CN-82K is a uniformly sampled version of CN-100k dataset which contains more unseen entities in the test split \cite{wang2021inductive}. \\

\noindent \textbf{ATOMIC}. ATOMIC \cite{sap2019atomic} contains rich social commonsense knowledge about day-to-day events. The dataset specifies the effects, needs, intents, and attributes of the actors in the events, covering nine relations and 877k knowledge tuples. Dataset built from ATOMIC for CSKG completion is first created in \citet{malaviya2020commonsense}. 

In our experiments, we follow \citet{wang2021inductive} to use CN-82K and ATOMIC. To better evaluate the generalizability of representation from MICO, we conduct experiments with the inductive splits in which one of entity nodes in a knowledge triplet from the valid and test split does not appear in the training dataset. Statistics of the converted sequence pairs from original datasets are shown in Table \ref{cskb:1}. 

\subsection{Evaluation Tasks}
Based on that CKGC and CQA can be unified into the same form of selecting alternatives given a premise, commonsense knowledge representations generated from MICO are evaluated on these two tasks.

\subsubsection{Zero-shot CQA}
The knowledge representation is evaluated on three multiple-choice CQA tasks: COPA \cite{roemmele2011choice}, SIQA \cite{sap2019social}, and CSQA \cite{talmor2019commonsenseqa}. Accuracy is used as the evaluation metric. For each task, the query composed by context and question can be converted into the form as a premise. The answers are viewed as possible plausible alternatives. In this way, the multiple-choice question can be solved by selecting the closet representation pairs generated from MICO given the query and candidate answers. We denote the representation for query as $\mathbf{q}$ and candidate answers as $\{\mathbf{a_1}, \mathbf{a_2}, ...\mathbf{a_m}\}$. The answer with the highest score is the predicted answer $i^*$, where
\begin{equation}
   i^* = \mathop{\arg\max}_{i = 1, ..., m} \textnormal{sim}(\mathbf{q}, \mathbf{a_i}).
\end{equation}
\\
\noindent \textbf{COPA}. Choice of Plausible Alternatives is a two-way multiple-choice commonsense reasoning task between events. COPA consists of 1,000 questions, 500 for the development set and 500 for the test set. To make the form of relation consistent with the training dataset in natural language, we substitute \textbf{cause} as \textbf{The cause for it was that} and \textbf{effect} as \textbf{As a result}. \\

\noindent \textbf{SIQA}. The queries in Social IQA are collected based on ATOMIC. Each question in the dataset describes social interactions and has three crowdsourced candidate answers. The dataset's development split and test split are used as zero-shot evaluation, containing 1,954 and 2,059 questions, respectively. \\

\noindent \textbf{CSQA}. The questions in CommonsenseQA are general questions about concepts in ConceptNet. Each question has five candidate answers. The development set is used as evaluation set, containing 1,221 questions.

\begin{table*}[t]
    \centering
    \small
    \begin{tabular}{lccccccc}
    \toprule
    \multicolumn{1}{c}{\multirow{2}{*}{Methods}} & \multicolumn{1}{c}{\multirow{2}{*}{Backbone}} & \multicolumn{1}{c}{\multirow{2}{*}{Knowledge Source}} & \multicolumn{2}{c}{COPA}  & \multicolumn{2}{c}{SIQA}  & CSQA   \\
    
    \multicolumn{1}{c}{}  & \multicolumn{1}{c}{} & \multicolumn{1}{c}{}  & \multicolumn{1}{c}{dev} & \multicolumn{1}{c}{test} & \multicolumn{1}{c}{dev} & \multicolumn{1}{c}{test} & dev \\
    \midrule
    Random & - & - & 50.0 & 50.0 & 33.3 & 33.3 & 25.0 \\
    RoBERTa-L & RoBERTa-L & - & 54.8 & 58.4 & 39.8 & 40.1 & 31.3 \\
    GPT2-L & GPT2-L & - & 62.4 & 63.6 & 42.8 & 43.3 & 40.4 \\
    self-talk & GPT2-[Distil/XL/L] & GPT2-[Distil/L/M] & 66.0 & - & 46.2 & 43.9 & 32.4 \\
    Dou & ALBERT-XXL-v2 & ALBERT-XXL-v2 & - & - & 44.1 & 42.0 & 50.9 \\
    \midrule
    KRL & RoBERTa-L & e.g., ATOMIC & - & - & 46.6 & 46.4 & 36.8 \\
    COMET-DynaGen & GPT2-M & COMET & - & - & 50.1 & 52.6 & - \\
    \midrule
    MICO & RoBERTa-L & ConceptNet & 73.2 & 75.2 & 44.6 & 45.4  & \textbf{51.0} \\
    MICO & RoBERTa-L & ATOMIC & \textbf{79.4} & \textbf{77.4} & \textbf{56.0} & \textbf{57.4} & 44.2 \\
    \bottomrule
    \end{tabular}
    \caption{Results on Zero-shot CQA tasks. COMET is the commonsense transformer trained on ATOMIC. For MICO, $k$ is set as 2. RoBERTa-L and GPT2-M have comparable parameter size. KRL is the knowledge representation method in KTL.}
    \label{ret:1}
    \vskip-1.5em
\end{table*}

\subsubsection{Inductive CKGC}
Inductive CKGC is an important task for CKG because unseen entity nodes are introduced in real-world CKGC from time to time and many distinct nodes may refer to same concept due to their free-form text description \cite{wang2021inductive}. In the inductive CKGC task, at least one of the nodes in knowledge triplets is not shown in the training dataset. Following \citet{wang2021inductive}, each triplet $(h,r,t)$ is measured in two directions: $(h,r, ?)$ and $(t, r^{-1}, ?)$. Inverse relations $r^{-1}$ are added as additional relation types. We use the MRR (mean reciprocal rank) and Hits@10 score as the evaluation metrics. 


\subsection{Baselines}
For zero-shot CQA tasks, we compare our method with baselines including pre-trained LMs (RoBERTa \cite{liu2019roberta}, GPT2 \cite{radford2019language}), using pre-trained LMs as knowledge sources (self-talk \cite{shwartz2020unsupervised}, Dou \cite{dou2022zero}), and pre-trained LMs trained on CKGs (KTL \cite{banerjee2020self}, COMET-DynaGen \cite{bosselut2021dynamic}).

For inductive CKGC tasks, we compare our method with ConvE \cite{dettmers2018convolutional}, RotatE \cite{sun2018rotate}, Malaviya \cite{malaviya2020commonsense} and InductivE \cite{wang2021inductive}. More details about baseline models for the two tasks are introduced in Appendix \ref{sec:appendix2}. 

\subsection{Implementation Details}
Our experiments are run on RTX A6000. Each experiment is run on a single GPU card. The training batch size is 196. Max sequence length for training is 32. The learning rate is set as 1e-5 for Bert-base and RoBERTa-base. For RoBERTa-large (RoBERTa-L), the learning rate is set as 5e-6.
We use AdamW \cite{loshchilov2018decoupled} optimizer. For experiments with the MICO framework, $\tau$ is set as 0.07. The valid set is evaluated by constrastive loss metric and used to select a best model for further evaluation. The models are trained for 10 epochs and early stopped when the change of validation loss is within 1\%.

\section{Results}


\subsection{Main Results}
The main results include MICO on the zero-shot CQA and the inductive CKGC.

\subsubsection{Results on Zero-shot CQA}
The results on CQA tasks are shown in Table \ref{ret:1}. Baseline systems based on pre-trained language models such as RoBERTa-L and GPT2-L provides strong baselines. Simply comparing the language model score from RoBERTa-L or GPT2-L outperforms random guess by a large margin. This shows that pre-trained language models are encoded with useful knowledge which can benefit the CQA tasks.

MICO generates knowledge representation encoded with commonsense knowledge by the finetuned LMs with self-supervision signal from CKGs. Compared with methods such as self-talk \cite{shwartz2020unsupervised} and Dou \cite{dou2022zero}, our method outperforms all the evaluation datasets. Self-talk and Dou utilize the pre-trained language models as knowledge source and mine relevant knowledge that may benefit the CQA tasks. However, such knowledge is still not sufficient. By fintuning on CKG, MICO can successfully inject the commonsense knowledge into pre-trained LMs and generate meaningful representation benefiting the CQA tasks.

MICO provides an efficient way to inject CKGs into pre-trained LMs. MICO solely trained on one knowledge source can achieve comparable performance or outperforms KRL \cite{banerjee2020self} and COMET-DynaGen \cite{bosselut2021dynamic}. KRL encodes the knowledge triplets into embeddings separately and then fuses two of them to predict the third one. Compared to KRL, MICO generates knowledge representations for sequence pairs, in which the relation interacts better with node entities as they are concatenated on the contextual level. COMET-DynaGen solves CQA tasks by utilizing the clarifications generated from COMET. However COMET is a generative model and always introduces novel entities \cite{wang2021inductive}, which may not be related to the query. Compared to COMET-DynaGen, MICO solves the CQA task by simply generating CKG related representations and comparing the similarity, also saving the cost of generating multi-step clarifications. 


    
Another finding is that the representation generated from MICO can be easily generalized to out-of-domain datasets. SIQA achieves best results when ATOMIC used as the knowledge source and CSQA achieves best results when ConceptNet used as the knowledge source. This is because SIQA is built based on ATOMIC and CSQA is built on ConceptNet. MICO still benefits the task for COPA, which requires commonsense knowledge but is not closely related to the two knowledge sources. This shows that the knowledge representation generated by MICO can generalize across tasks.

\begin{table}[t]
  \centering
  \small
  \begin{tabular}{lcccc}
  \toprule
  \multicolumn{1}{c}{\multirow{2}{*}{Model}} & \multicolumn{2}{c}{ConceptNet}  & \multicolumn{2}{c}{ATOMIC} \\
  \multicolumn{1}{c}{}  & \multicolumn{1}{c}{MRR} & \multicolumn{1}{c}{Hits@10} & \multicolumn{1}{c}{MRR} & \multicolumn{1}{c}{Hits@10} \\
  \midrule
  ConvE & 0.21 & 0.40 & 0.08 & 0.09 \\
  RotatE & 0.32 & 0.50 & 0.10 & 0.12 \\
  Malaviya  & 12.29 & 19.36 & 0.02 & 0.07 \\
  InductivE & \textbf{18.15} & \textbf{29.37} & 2.51 & 5.45 \\
  \midrule
  MICO$^{\ast}$ & 9.00 & 19.06 & 7.07 & 13.52 \\
  MICO$^{\diamondsuit}$ & 9.08 & 18.73 & 7.52 & 14.46 \\
  MICO$^{\heartsuit}$ & 10.92 & 22.07 & \textbf{8.13} & \textbf{15.69} \\
  \bottomrule
  \end{tabular}
  \caption{Results on inductive CKGC. $\textnormal{MICO}^{\ast}$ for BERT-base, $\textnormal{MICO}^{\diamondsuit}$ for RoBERTa-base, $\textnormal{MICO}^{\heartsuit}$ for RoBERTa-L. $k$ is set as 2.}
  \label{ret:2}
\end{table}

\subsubsection{Results on Inductive CKGC}
MICO enhances the commonsense representation by the contrast information between knowledge triplets and can generalize to unseen entity nodes. Results on the inductive CKGC task are shown in Table \ref{ret:2}. Previous methods such as ConvE \cite{dettmers2018convolutional} and RotatE \cite{sun2018rotate} rely on relation link between entities to learn entity embedding. These methods perform bad when new entities come with no link to previous nodes existing. Methods such as Malaviya \cite{malaviya2020commonsense} or InductivE \cite{wang2021inductive} apply pre-trained LMs to initialize the node embedding and then focus on utilizing subgraph structure to improve the generalizability of node features by GCN. However, the CKG is sparse and the average degree for each node is roughly around 2 for both CKGs. Thus MICO focuses on learning the context information of node entities and achieves better performance on ATOMIC while comparable on ConceptNet.

MICO achieves better performance than InductivE on ATOMIC while otherwise on ConceptNet. The entity nodes contain 3.93 words on average in ConceptNet and 6.12 words on average in ATOMIC. MICO encodes the node textual description by pre-trained LMs and longer word sequences results in better distinguishable node feature. This may explains why MICO performs better on ATOMIC than on ConceptNet compared to InductivE. InductivE relies on learning the neighboring graph structure by GCN. However in ATOMIC, the entity nodes are more complex than those in ConceptNet so capturing the graph structure is not enough to learn good commonsense representation.

\subsection{Ablation Study and Analysis}
In this part, we analyze the influence of backbone models, number of candidate positive tails $k$, and hard positive selection in MICO. For evaluation on CQA tasks, the results are reported on the development set of SIQA and CSQA, and combination of development set and test set of COPA. 

\subsubsection{Backbone Pre-trained LMs}
\begin{table}[t]
    \small
    \centering
    \begin{tabular}{p{2cm}cccc}
    \toprule
    Backbone    & CKG & COPA   & SIQA  & CSQA \\ \midrule
    \multirow{3}{*}{BERT-base} & -  & 45.9  & 37.1 & 21.5  \\
                            & ConceptNet  & 65.2  & 39.1  & \underline{42.9} \\
                            & ATOMIC  & \underline{71.3}  & \underline{48.9} & 40.7 \\ \midrule
    \multirow{3}{*}{RoBERTa-base} & -  & 53.5 & 38.4 & 29.2 \\
                            & ConceptNet  & 67.7 & 39.8  & \underline{44.7} \\
                            & ATOMIC  & \underline{72.0} & \underline{51.9} & 40.8  \\
                            \midrule
    \multirow{3}{*}{RoBERTa-L} & -  & 56.6 & 39.8 & 31.3 \\
                            & ConceptNet  & 74.2 & 44.6  & \underline{51.0} \\
                            & ATOMIC  & \underline{78.4} & \underline{56.0} & 44.2  \\ 
    \bottomrule 
    \end{tabular}
    \caption{Backbone model study on two CKGs and evaluation on CQA tasks. For MICO, $k$ is set as 2 during training.}
    \label{ret:5}
\end{table}

The results on different backbone models are shown in Table \ref{ret:5}. MICO trained on different backbone models show consistent pattern on the three commonsense QA tasks. First, MICO trained with CKGs outperform baseline models without any CKG knowledge. Second, MICO trained with ConceptNet achieves better performance on CSQA and trained with ATOMIC achieves better performance on COPA and SIQA. 

\subsubsection{Hyper-parameter $k$}
In this part, we study how the number of positive tails $k$ influence the effects of MICO. For simplicity, we study the influence of $k$ on two graphs with BERT-base as the backbone model. 
\begin{table}[t]
    \small
    \centering

\begin{tabular}{ccccc}
\toprule
CKG    & \multicolumn{1}{l}{$k$} & COPA   & SIQA  & CSQA \\ \midrule
\multirow{4}{*}{ConceptNet} & 1  & 64.9  & 39.3  & 42.1  \\
                            & 2  & \underline{65.2}  & 39.1  & 42.9 \\
                            & 3  & 64.7  & 39.2 & 42.8 \\
                            & 4  & 64.0 & \underline{39.8} & \underline{\textbf{43.9}} \\ \midrule
\multirow{4}{*}{ATOMIC}     & 1  & \underline{\textbf{72.2}} & 48.2 & 40.7 \\
                            & 2  & 71.3 & 48.9  & 40.7 \\
                            & 3  & 72.1 & 48.9 & \underline{41.0}  \\
                            & 4  & 70.2 & \underline{\textbf{49.2}} & 40.5 \\ \bottomrule
\end{tabular}
    \caption{Hyper-parameter study of $k$ on two CKGs and evaluation on zero-shot CQA tasks.}
    \label{ret:7}
\end{table}

\begin{table}[t]
  \centering
  \small
  \begin{tabular}{lllll}
  \toprule
  \multicolumn{1}{c}{\multirow{2}{*}{$k$}} & \multicolumn{2}{c}{ConceptNet}  & \multicolumn{2}{c}{ATOMIC} \\
  \multicolumn{1}{c}{}  & \multicolumn{1}{c}{MRR} & \multicolumn{1}{c}{Hits@10} & \multicolumn{1}{c}{MRR} & \multicolumn{1}{c}{Hits@10} \\
    \midrule
    1 & 8.81 & 18.82 & 6.69 & 12.78 \\
    2 & 9.00 & 19.06 & 7.07 & 13.52 \\
    3 & \textbf{9.41} & \textbf{19.65} & 7.08 & 13.46 \\
    4 & 9.21 & 19.22 & \textbf{7.14} & \textbf{13.58} \\
    \bottomrule
    \end{tabular}
    \caption{Hyper-parameter study of $k$ on two CKGs and evaluation on inductive CKGC tasks. }
    \label{ret:8}
\end{table}

The performance of CQA tasks and inductive CKGC tasks under the influence of $k$ is shown in Table \ref{ret:7} and Table \ref{ret:8}. MICO generally performs better on CSQA when trained on ConceptNet and SIQA when trained on ATOMIC. This is because the questions in each task are more related to the knowledge in the corresponding CKG. 


The performances on the inductive CKGC mostly increase as $k$ increases. This indicates that larger $k$ helps the model generalize better in pairing the in-domain knowledge sequences. However for ConceptNet, the performance drops when $k$ is greater than 3. The limited average degree of nodes in ConceptNet may explain this as larger $k$ does not induce new candidate tails. Therefore, the model tends to fit the seen nodes better. 

\subsubsection{Sampling Strategy}
We analyze the influence of selecting a hard positive compared with randomly sampling a positive from candidate sets. The results are show in Table \ref{ret:6}. The experiments are conducted on the backbone model with BERT-base and $k=2$. Compared to random sampling, MICO mostly outperforms on the three datasets. This indicates that the hard positive during training can benefit the generalization of the representation. The only exception is training on ATOMIC and testing on SIQA. The possible explanation is that there is possibly a distribution gap between the training dataset and SIQA dataset. However, generally our sampling strategy can improve the generalizability of the representation.

\begin{table}[t]
    \small
    \centering
    \begin{tabular}{ccccc}
    \toprule
    CKG    & Sampling & COPA   & SIQA  & CSQA \\ 
    \midrule
    \multirow{2}{*}{ConceptNet} & Random  & 64.0  & 38.3  & 41.3  \\
                            & MICO  & \underline{65.2}  & \underline{39.1} & \underline{42.9} \\
                            \midrule
    \multirow{2}{*}{ATOMIC} & Random  & 71.1 & \underline{49.2} & 40.3 \\
                            & MICO  & \underline{71.3} & 48.9  & \underline{40.7} \\
    \bottomrule 
    \end{tabular}
    \caption{Ablation study on sampling strategy on two CKGs and evaluation on CQA tasks. $k$ is set as 2 during training.}
    \label{ret:6}
\end{table}

\section{Discussion}

This section shows that transformers from MICO can construct commonsense representations for CKG and benefit commonsense knowledge retrieval given queries. One example from SIQA and retrieved possible alternatives from ATOMIC is shown in Table \ref{ret:8}. We first encode all the alternatives in CKG by the transformer finetuned with ATOMIC and original transformer without any finetuning. The top 5 retrieved nodes are listed by the ranks of similarity score in descending order. 

We can find that the transformer finetuned on CKG can successfully pair the query with reasonable alternatives from CKG compared to original pre-trained transformer. Therefore, our method provides an efficient way to collect the related knowledge from CKG and may benefit the researches which require retrieved implicit background knowledge to reason over. 

However, the representation generated from MICO still has some drawbacks as shown in the results. ``Jordan look for it at library'' would be a reasonable node instead of ``Jordan look for it at home''. This shows that the representation still need future work to distinguish the detailed concepts.

\begin{table}[t]
  \centering
  \small
  \begin{tabular}{lp{6cm}}
  \toprule
  \multirow{2}{*}{Query} & Jordan left their book if the library after \\
  & studying all day. As a result, Jordan wanted to \\
  \midrule
   & \textbf{to go back and get the book}  \\     \cline{2-2}
  w/  & to look for it at home \\ \cline{2-2}
  CKG  & \textbf{to check the lost and the found} \\ \cline{2-2}
    & \textbf{to pick up the item they forgot} \\ \cline{2-2}
    & \textbf{to try to remember where they put it} \\ 
  \midrule
  \multirow{2}{*}{} & take a pet along to the apartment viewing \\
  & and scares the landlord \\ \cline{2-2}
    & seen \\ \cline{2-2}
  w/o  & resolute \\ \cline{2-2}
  CKG  & to contemplate circumstances and possible outcomes \\ \cline{2-2}
    & to go out and form a relationship \\
  \bottomrule
  \end{tabular}
  \caption{Comparison of retrieved alternatives from representations extracted RoBERTa-L with CKG (ATOMIC) and without CKG on question from SIQA task. Reasonable alternatives are in boldface.}
  \label{ret:9}
\end{table}

\section{Conclusion}

In this paper, we propose a MICO, a multi-alternative contrastive learning framework over commonsense knowledge graphs to learn commonsense knowledge representation. The framework converts the knowledge triplets into sequence pairs and learns superior knowledge representation through contrastive learning techniques. The generated representations perform well over zero-shot CQA tasks and inductive CKGC tasks. Furthermore, for CQA tasks, the related knowledge can be provided by simply retrieving the commonsense knowledge representations of CKGs. 

\section*{Acknowledgements}
The authors of this paper were supported by the NSFC Fund (U20B2053) from the NSFC of China, the RIF (R6020-19 and R6021-20) and the GRF (16211520) from RGC of Hong Kong, the MHKJFS (MHP/001/19) from ITC of Hong Kong and the National Key R\&D Program of China (2019YFE0198200) with special thanks to HKMAAC and CUSBLT, and  the Jiangsu Province Science and Technology Collaboration Fund (BZ2021065). We also thank the support from the UGC Research Matching Grants (RMGS20EG01-D, RMGS20CR11, RMGS20CR12, RMGS20EG19, RMGS20EG21).

\section*{Limitations}
As shown in the discussion, the commonsense knowledge representation generated from MICO can capture the rough meanings of the whole word sequences. While for detailed concepts, the representation failed to distinguish since the representation is extracted from a specific token to represent the meaning of the whole word sequence. However, concepts are key elements in semantics so future work is still needed to improve the representation.

\bibliography{anthology,custom}

\begin{thebibliography}{51}
\expandafter\ifx\csname natexlab\endcsname\relax\def\natexlab#1{#1}\fi

\bibitem[{Bachman et~al.(2019)Bachman, Hjelm, and
  Buchwalter}]{bachman2019learning}
Philip Bachman, R~Devon Hjelm, and William Buchwalter. 2019.
\newblock Learning representations by maximizing mutual information across
  views.
\newblock \emph{Advances in neural information processing systems}, 32.

\bibitem[{Banerjee and Baral(2020)}]{banerjee2020self}
Pratyay Banerjee and Chitta Baral. 2020.
\newblock Self-supervised knowledge triplet learning for zero-shot question
  answering.
\newblock In \emph{Proceedings of the 2020 Conference on Empirical Methods in
  Natural Language Processing (EMNLP)}, pages 151--162.

\bibitem[{Bisk et~al.(2020)Bisk, Zellers, Gao, Choi et~al.}]{bisk2020piqa}
Yonatan Bisk, Rowan Zellers, Jianfeng Gao, Yejin Choi, et~al. 2020.
\newblock Piqa: Reasoning about physical commonsense in natural language.
\newblock In \emph{Proceedings of the AAAI Conference on Artificial
  Intelligence}, volume~34, pages 7432--7439.

\bibitem[{Bollacker et~al.(2008)Bollacker, Evans, Paritosh, Sturge, and
  Taylor}]{bollacker2008freebase}
Kurt Bollacker, Colin Evans, Praveen Paritosh, Tim Sturge, and Jamie Taylor.
  2008.
\newblock Freebase: a collaboratively created graph database for structuring
  human knowledge.
\newblock In \emph{Proceedings of the 2008 ACM SIGMOD international conference
  on Management of data}, pages 1247--1250.

\bibitem[{Bordes et~al.(2013)Bordes, Usunier, Garcia-Duran, Weston, and
  Yakhnenko}]{bordes2013translating}
Antoine Bordes, Nicolas Usunier, Alberto Garcia-Duran, Jason Weston, and Oksana
  Yakhnenko. 2013.
\newblock Translating embeddings for modeling multi-relational data.
\newblock \emph{Advances in neural information processing systems}, 26.

\bibitem[{Bosselut et~al.(2021)Bosselut, Le~Bras, and
  Choi}]{bosselut2021dynamic}
Antoine Bosselut, Ronan Le~Bras, and Yejin Choi. 2021.
\newblock Dynamic neuro-symbolic knowledge graph construction for zero-shot
  commonsense question answering.
\newblock In \emph{Proceedings of the 35th AAAI Conference on Artificial
  Intelligence (AAAI)}.

\bibitem[{Bosselut et~al.(2019)Bosselut, Rashkin, Sap, Malaviya, Celikyilmaz,
  and Choi}]{bosselut2019comet}
Antoine Bosselut, Hannah Rashkin, Maarten Sap, Chaitanya Malaviya, Asli
  Celikyilmaz, and Yejin Choi. 2019.
\newblock Comet: Commonsense transformers for automatic knowledge graph
  construction.
\newblock In \emph{Proceedings of the 57th Annual Meeting of the Association
  for Computational Linguistics}, pages 4762--4779.

\bibitem[{Carlsson et~al.(2020)Carlsson, Gyllensten, Gogoulou, Hellqvist, and
  Sahlgren}]{carlsson2020semantic}
Fredrik Carlsson, Amaru~Cuba Gyllensten, Evangelia Gogoulou,
  Erik~Ylip{\"a}{\"a} Hellqvist, and Magnus Sahlgren. 2020.
\newblock Semantic re-tuning with contrastive tension.
\newblock In \emph{International Conference on Learning Representations}.

\bibitem[{Dettmers et~al.(2018)Dettmers, Minervini, Stenetorp, and
  Riedel}]{dettmers2018convolutional}
Tim Dettmers, Pasquale Minervini, Pontus Stenetorp, and Sebastian Riedel. 2018.
\newblock Convolutional 2d knowledge graph embeddings.
\newblock In \emph{Proceedings of the AAAI Conference on Artificial
  Intelligence}, volume~32.

\bibitem[{Dou and Peng(2022)}]{dou2022zero}
Zi-Yi Dou and Nanyun Peng. 2022.
\newblock Zero-shot commonsense question answering with cloze translation and
  consistency optimization.
\newblock \emph{AAAI}.

\bibitem[{Fang et~al.(2021)Fang, Zhang, Wang, Song, and He}]{fang2021discos}
Tianqing Fang, Hongming Zhang, Weiqi Wang, Yangqiu Song, and Bin He. 2021.
\newblock Discos: Bridging the gap between discourse knowledge and commonsense
  knowledge.
\newblock In \emph{Proceedings of the Web Conference 2021}, pages 2648--2659.

\bibitem[{Feng et~al.(2020)Feng, Chen, Lin, Wang, Yan, and
  Ren}]{feng2020scalable}
Yanlin Feng, Xinyue Chen, Bill~Yuchen Lin, Peifeng Wang, Jun Yan, and Xiang
  Ren. 2020.
\newblock Scalable multi-hop relational reasoning for knowledge-aware question
  answering.
\newblock In \emph{Proceedings of the 2020 Conference on Empirical Methods in
  Natural Language Processing (EMNLP)}, pages 1295--1309.

\bibitem[{Gao et~al.(2021)Gao, Yao, and Chen}]{gao2021simcse}
Tianyu Gao, Xingcheng Yao, and Danqi Chen. 2021.
\newblock Simcse: Simple contrastive learning of sentence embeddings.
\newblock In \emph{Proceedings of the 2021 Conference on Empirical Methods in
  Natural Language Processing}, pages 6894--6910.

\bibitem[{He et~al.(2020)He, Fan, Wu, Xie, and Girshick}]{he2020momentum}
Kaiming He, Haoqi Fan, Yuxin Wu, Saining Xie, and Ross Girshick. 2020.
\newblock Momentum contrast for unsupervised visual representation learning.
\newblock In \emph{Proceedings of the IEEE/CVF Conference on Computer Vision
  and Pattern Recognition}, pages 9729--9738.

\bibitem[{Hwang et~al.(2021)Hwang, Bhagavatula, Le~Bras, Da, Sakaguchi,
  Bosselut, and Choi}]{hwang2021comet}
Jena~D Hwang, Chandra Bhagavatula, Ronan Le~Bras, Jeff Da, Keisuke Sakaguchi,
  Antoine Bosselut, and Yejin Choi. 2021.
\newblock (comet-) atomic 2020: On symbolic and neural commonsense knowledge
  graphs.
\newblock In \emph{Proceedings of the AAAI Conference on Artificial
  Intelligence}, volume~35, pages 6384--6392.

\bibitem[{Kenton and Toutanova(2019)}]{kenton2019bert}
Jacob Devlin Ming-Wei~Chang Kenton and Lee~Kristina Toutanova. 2019.
\newblock Bert: Pre-training of deep bidirectional transformers for language
  understanding.
\newblock In \emph{Proceedings of NAACL-HLT}, pages 4171--4186.

\bibitem[{Kipf and Welling(2017)}]{kipf2017semi}
Thomas~N Kipf and Max Welling. 2017.
\newblock Semi-supervised classification with graph convolutional networks.
\newblock \emph{International Conference on Learning Representations (ICLR)}.

\bibitem[{Li et~al.(2016)Li, Taheri, Tu, and Gimpel}]{li2016commonsense}
Xiang Li, Aynaz Taheri, Lifu Tu, and Kevin Gimpel. 2016.
\newblock Commonsense knowledge base completion.
\newblock In \emph{Proceedings of the 54th Annual Meeting of the Association
  for Computational Linguistics (Volume 1: Long Papers)}, pages 1445--1455.

\bibitem[{Lin et~al.(2019)Lin, Chen, Chen, and Ren}]{lin2019kagnet}
Bill~Yuchen Lin, Xinyue Chen, Jamin Chen, and Xiang Ren. 2019.
\newblock Kagnet: Knowledge-aware graph networks for commonsense reasoning.
\newblock In \emph{Proceedings of the 2019 Conference on Empirical Methods in
  Natural Language Processing and the 9th International Joint Conference on
  Natural Language Processing (EMNLP-IJCNLP)}, pages 2829--2839.

\bibitem[{Liu and Singh(2004)}]{liu2004conceptnet}
Hugo Liu and Push Singh. 2004.
\newblock Conceptnet—a practical commonsense reasoning tool-kit.
\newblock \emph{BT technology journal}, 22(4):211--226.

\bibitem[{Liu et~al.(2019)Liu, Ott, Goyal, Du, Joshi, Chen, Levy, Lewis,
  Zettlemoyer, and Stoyanov}]{liu2019roberta}
Yinhan Liu, Myle Ott, Naman Goyal, Jingfei Du, Mandar Joshi, Danqi Chen, Omer
  Levy, Mike Lewis, Luke Zettlemoyer, and Veselin Stoyanov. 2019.
\newblock Roberta: A robustly optimized bert pretraining approach.
\newblock \emph{arXiv preprint arXiv:1907.11692}.

\bibitem[{Loshchilov and Hutter(2018)}]{loshchilov2018decoupled}
Ilya Loshchilov and Frank Hutter. 2018.
\newblock Decoupled weight decay regularization.
\newblock In \emph{International Conference on Learning Representations}.

\bibitem[{Lv et~al.(2020)Lv, Guo, Xu, Tang, Duan, Gong, Shou, Jiang, Cao, and
  Hu}]{lv2020graph}
Shangwen Lv, Daya Guo, Jingjing Xu, Duyu Tang, Nan Duan, Ming Gong, Linjun
  Shou, Daxin Jiang, Guihong Cao, and Songlin Hu. 2020.
\newblock Graph-based reasoning over heterogeneous external knowledge for
  commonsense question answering.
\newblock In \emph{Proceedings of the AAAI Conference on Artificial
  Intelligence}, volume~34, pages 8449--8456.

\bibitem[{Ma et~al.(2021)Ma, Ilievski, Francis, Bisk, Nyberg, and
  Oltramari}]{ma2021knowledge}
Kaixin Ma, Filip Ilievski, Jonathan Francis, Yonatan Bisk, Eric Nyberg, and
  Alessandro Oltramari. 2021.
\newblock Knowledge-driven data construction for zero-shot evaluation in
  commonsense question answering.
\newblock In \emph{35th AAAI Conference on Artificial Intelligence}.

\bibitem[{Malaviya et~al.(2020)Malaviya, Bhagavatula, Bosselut, and
  Choi}]{malaviya2020commonsense}
Chaitanya Malaviya, Chandra Bhagavatula, Antoine Bosselut, and Yejin Choi.
  2020.
\newblock Commonsense knowledge base completion with structural and semantic
  context.
\newblock In \emph{Proceedings of the AAAI conference on artificial
  intelligence}, volume~34, pages 2925--2933.

\bibitem[{Miller(1995)}]{miller1995wordnet}
George~A Miller. 1995.
\newblock Wordnet: a lexical database for english.
\newblock \emph{Communications of the ACM}, 38(11):39--41.

\bibitem[{Nickel et~al.(2011)Nickel, Tresp, and Kriegel}]{nickel2011three}
Maximilian Nickel, Volker Tresp, and Hans-Peter Kriegel. 2011.
\newblock A three-way model for collective learning on multi-relational data.
\newblock In \emph{Icml}.

\bibitem[{Niu et~al.(2022)Niu, Li, Zhang, and Pu}]{niu2022cake}
Guanglin Niu, Bo~Li, Yongfei Zhang, and Shiliang Pu. 2022.
\newblock Cake: A scalable commonsense-aware framework for multi-view knowledge
  graph completion.
\newblock In \emph{Proceedings of the 60th Annual Meeting of the Association
  for Computational Linguistics (Volume 1: Long Papers)}, pages 2867--2877.

\bibitem[{Paul and Frank(2019)}]{paul2019ranking}
Debjit Paul and Anette Frank. 2019.
\newblock Ranking and selecting multi-hop knowledge paths to better predict
  human needs.
\newblock In \emph{Proceedings of the 2019 Conference of the North American
  Chapter of the Association for Computational Linguistics: Human Language
  Technologies, Volume 1 (Long and Short Papers)}, pages 3671--3681.

\bibitem[{Radford et~al.(2019)Radford, Wu, Child, Luan, Amodei, Sutskever
  et~al.}]{radford2019language}
Alec Radford, Jeffrey Wu, Rewon Child, David Luan, Dario Amodei, Ilya
  Sutskever, et~al. 2019.
\newblock Language models are unsupervised multitask learners.
\newblock \emph{OpenAI blog}, 1(8):9.

\bibitem[{Roemmele et~al.(2011)Roemmele, Bejan, and
  Gordon}]{roemmele2011choice}
Melissa Roemmele, Cosmin~Adrian Bejan, and Andrew~S Gordon. 2011.
\newblock Choice of plausible alternatives: An evaluation of commonsense causal
  reasoning.
\newblock In \emph{2011 AAAI Spring Symposium Series}.

\bibitem[{Sap et~al.(2019{\natexlab{a}})Sap, Le~Bras, Allaway, Bhagavatula,
  Lourie, Rashkin, Roof, Smith, and Choi}]{sap2019atomic}
Maarten Sap, Ronan Le~Bras, Emily Allaway, Chandra Bhagavatula, Nicholas
  Lourie, Hannah Rashkin, Brendan Roof, Noah~A Smith, and Yejin Choi.
  2019{\natexlab{a}}.
\newblock Atomic: An atlas of machine commonsense for if-then reasoning.
\newblock In \emph{Proceedings of the AAAI Conference on Artificial
  Intelligence}, volume~33, pages 3027--3035.

\bibitem[{Sap et~al.(2019{\natexlab{b}})Sap, Rashkin, Chen, Le~Bras, and
  Choi}]{sap2019social}
Maarten Sap, Hannah Rashkin, Derek Chen, Ronan Le~Bras, and Yejin Choi.
  2019{\natexlab{b}}.
\newblock Social iqa: Commonsense reasoning about social interactions.
\newblock In \emph{Proceedings of the 2019 Conference on Empirical Methods in
  Natural Language Processing and the 9th International Joint Conference on
  Natural Language Processing (EMNLP-IJCNLP)}, pages 4463--4473.

\bibitem[{Shang et~al.(2019)Shang, Tang, Huang, Bi, He, and
  Zhou}]{shang2019end}
Chao Shang, Yun Tang, Jing Huang, Jinbo Bi, Xiaodong He, and Bowen Zhou. 2019.
\newblock End-to-end structure-aware convolutional networks for knowledge base
  completion.
\newblock In \emph{Proceedings of the AAAI Conference on Artificial
  Intelligence}, volume~33, pages 3060--3067.

\bibitem[{Shwartz et~al.(2020)Shwartz, West, Le~Bras, Bhagavatula, and
  Choi}]{shwartz2020unsupervised}
Vered Shwartz, Peter West, Ronan Le~Bras, Chandra Bhagavatula, and Yejin Choi.
  2020.
\newblock Unsupervised commonsense question answering with self-talk.
\newblock In \emph{Proceedings of the 2020 Conference on Empirical Methods in
  Natural Language Processing (EMNLP)}, pages 4615--4629.

\bibitem[{Singh et~al.(2002)}]{singh2002public}
Push Singh et~al. 2002.
\newblock The public acquisition of commonsense knowledge.
\newblock In \emph{Proceedings of AAAI Spring Symposium: Acquiring (and Using)
  Linguistic (and World) Knowledge for Information Access}.

\bibitem[{Speer et~al.(2017)Speer, Chin, and Havasi}]{speer2017conceptnet}
Robyn Speer, Joshua Chin, and Catherine Havasi. 2017.
\newblock Conceptnet 5.5: An open multilingual graph of general knowledge.
\newblock In \emph{Thirty-first AAAI conference on artificial intelligence}.

\bibitem[{Sun et~al.(2018)Sun, Deng, Nie, and Tang}]{sun2018rotate}
Zhiqing Sun, Zhi-Hong Deng, Jian-Yun Nie, and Jian Tang. 2018.
\newblock Rotate: Knowledge graph embedding by relational rotation in complex
  space.
\newblock In \emph{International Conference on Learning Representations}.

\bibitem[{Sun et~al.(2019)Sun, Deng, Nie, and Tang}]{sun2019rotate}
Zhiqing Sun, Zhi-Hong Deng, Jian-Yun Nie, and Jian Tang. 2019.
\newblock Rotate: Knowledge graph embedding by relational rotation in complex
  space.
\newblock \emph{arXiv preprint arXiv:1902.10197}.

\bibitem[{Talmor et~al.(2019)Talmor, Herzig, Lourie, and
  Berant}]{talmor2019commonsenseqa}
Alon Talmor, Jonathan Herzig, Nicholas Lourie, and Jonathan Berant. 2019.
\newblock Commonsenseqa: A question answering challenge targeting commonsense
  knowledge.
\newblock In \emph{Proceedings of NAACL-HLT}, pages 4149--4158.

\bibitem[{Tian et~al.(2020)Tian, Krishnan, and Isola}]{tian2020contrastive}
Yonglong Tian, Dilip Krishnan, and Phillip Isola. 2020.
\newblock Contrastive multiview coding.
\newblock In \emph{European conference on computer vision}, pages 776--794.
  Springer.

\bibitem[{Toutanova et~al.(2015)Toutanova, Chen, Pantel, Poon, Choudhury, and
  Gamon}]{toutanova2015representing}
Kristina Toutanova, Danqi Chen, Patrick Pantel, Hoifung Poon, Pallavi
  Choudhury, and Michael Gamon. 2015.
\newblock Representing text for joint embedding of text and knowledge bases.
\newblock In \emph{Proceedings of the 2015 conference on empirical methods in
  natural language processing}, pages 1499--1509.

\bibitem[{Trouillon et~al.(2016)Trouillon, Welbl, Riedel, Gaussier, and
  Bouchard}]{trouillon2016complex}
Th{\'e}o Trouillon, Johannes Welbl, Sebastian Riedel, {\'E}ric Gaussier, and
  Guillaume Bouchard. 2016.
\newblock Complex embeddings for simple link prediction.
\newblock In \emph{International conference on machine learning}, pages
  2071--2080. PMLR.

\bibitem[{Wang et~al.(2021)Wang, Wang, Huang, You, Leskovec, and
  Kuo}]{wang2021inductive}
Bin Wang, Guangtao Wang, Jing Huang, Jiaxuan You, Jure Leskovec, and C-C~Jay
  Kuo. 2021.
\newblock Inductive learning on commonsense knowledge graph completion.
\newblock In \emph{2021 International Joint Conference on Neural Networks
  (IJCNN)}, pages 1--8. IEEE.

\bibitem[{Wang et~al.(2014)Wang, Zhang, Feng, and Chen}]{wang2014knowledge}
Zhen Wang, Jianwen Zhang, Jianlin Feng, and Zheng Chen. 2014.
\newblock Knowledge graph embedding by translating on hyperplanes.
\newblock In \emph{Proceedings of the AAAI Conference on Artificial
  Intelligence}, volume~28.

\bibitem[{Xu et~al.(2021)Xu, Zhu, Xu, Liu, Zeng, and Huang}]{xu2021fusing}
Yichong Xu, Chenguang Zhu, Ruochen Xu, Yang Liu, Michael Zeng, and Xuedong
  Huang. 2021.
\newblock Fusing context into knowledge graph for commonsense question
  answering.
\newblock In \emph{Findings of the Association for Computational Linguistics:
  ACL-IJCNLP 2021}, pages 1201--1207.

\bibitem[{Yang et~al.(2019)Yang, Wang, Liu, Liu, Lyu, Wu, She, and
  Li}]{yang2019enhancing}
An~Yang, Quan Wang, Jing Liu, Kai Liu, Yajuan Lyu, Hua Wu, Qiaoqiao She, and
  Sujian Li. 2019.
\newblock Enhancing pre-trained language representations with rich knowledge
  for machine reading comprehension.
\newblock In \emph{Proceedings of the 57th Annual Meeting of the Association
  for Computational Linguistics}, pages 2346--2357.

\bibitem[{Yao et~al.(2019)Yao, Mao, and Luo}]{yao2019kg}
Liang Yao, Chengsheng Mao, and Yuan Luo. 2019.
\newblock Kg-bert: Bert for knowledge graph completion.
\newblock \emph{arXiv preprint arXiv:1909.03193}.

\bibitem[{Yasunaga et~al.(2021)Yasunaga, Ren, Bosselut, Liang, and
  Leskovec}]{yasunaga2021qa}
Michihiro Yasunaga, Hongyu Ren, Antoine Bosselut, Percy Liang, and Jure
  Leskovec. 2021.
\newblock Qa-gnn: Reasoning with language models and knowledge graphs for
  question answering.
\newblock In \emph{Proceedings of the 2021 Conference of the North American
  Chapter of the Association for Computational Linguistics: Human Language
  Technologies}, pages 535--546.

\bibitem[{Zellers et~al.(2018)Zellers, Bisk, Schwartz, and
  Choi}]{zellers2018swag}
Rowan Zellers, Yonatan Bisk, Roy Schwartz, and Yejin Choi. 2018.
\newblock Swag: A large-scale adversarial dataset for grounded commonsense
  inference.
\newblock In \emph{EMNLP}.

\bibitem[{Zhang et~al.(2021)Zhang, Bosselut, Yasunaga, Ren, Liang, Manning, and
  Leskovec}]{zhang2021greaselm}
Xikun Zhang, Antoine Bosselut, Michihiro Yasunaga, Hongyu Ren, Percy Liang,
  Christopher~D Manning, and Jure Leskovec. 2021.
\newblock Greaselm: Graph reasoning enhanced language models.
\newblock In \emph{International Conference on Learning Representations}.

\end{thebibliography}
\bibliographystyle{acl_natbib}

\appendix
\clearpage
\newpage

\section{Appendix}

\subsection{Templates for Relation}
\label{sec:appendix1}
For training dataset of two CSKGs, we used the version from InductivE \footnote{https://github.com/BinWang28/InductivE}.

\subsubsection{ATOMIC}
In ATOMIC, there are nine relations. The substitute template of original relations and reverse relations are shown in Table \ref{append:1}. \textit{PersonX} and \textit{PersonY} are substitued by "John" or "Tom" respectively.

\begin{table}[t]
    \centering
    \small
    \begin{tabular}{lc}
    \toprule
    Relation & rel template \\
    \midrule
    xAttr & PersonX is seen as \\
    xEffect & as a result, PersonX will \\
    xWant & as a result, PersonX wants \\
    xNeed & but before, PersonX needed \\
    xReact & as a result, PersonX feels \\
    xIntent & because PersonX wanted \\
    oEffect & as a result, PersonY or others will \\
    oReact & as a result, PersonY or others feel \\
    oWant & as a result, PersonY or others want \\
    \midrule
    xAttr rev & "PersonX is seen as", "because PersonX" \\
    xEffect rev & "PersonX will", "because PersonX" \\
    xWant rev & "PersonX wants", "because PersonX" \\
    xNeed rev & "PersonX needs", "as a result PersonX" \\
    xReact rev & "PersonX feels", "because PersonX" \\
    xIntent rev & "PersonX wanted", "as a result PersonX" \\
    oEffect rev & "PersonY or others will", "because PersonX" \\
    oReact rev & "PersonY or others feel", "because PersonX" \\
    oWant rev & "PersonY or others want", "because PersonX" \\
    \bottomrule
    \end{tabular}
    \caption{Relation types and relation substitute templates from ATOMIC. \textit{rev} mean reverse relation. }
    \label{append:1}
\end{table}

\subsubsection{ConceptNet}

ConceptNet contains 34 relations, The substitute template of original relations and reverse relations are shown in Table \ref{append:2}.

\begin{table}[t]
    \centering
    \small
    \begin{tabular}{lc}
    \toprule
    Relation & relation templates \\
    \midrule
     AtLocation & located or found at or in or on \\
     CapableOf & is or are capable of \\
     NotCapableOf & is not or are not capable of \\
     Causes & causes \\
     CausesDesire & makes someone want \\
     CreatedBy &  is created by \\
     DefinedAs & is defined as \\
     DesireOf & desires \\
     Desires & desires \\
     NotDesires & do not desire \\
     HasA & has, possesses, or contains \\
     HasFirstSubevent & begins with the event or action \\
     HasLastSubevent & ends with the event or action \\
     HasPrerequisite & to do this, one requires \\
     HasProperty & can be characterized by being or having \\
     InstanceOf & is an example or instance of \\
     IsA & is a \\
     MadeOf & is made of \\
     MotivatedByGoal & is a step towards accomplishing the goal \\
     PartOf & is a part of \\
     ReceivesAction & can receive or be affected by the action \\
     SymbolOf & is a symbol of \\
     UsedFor & used for \\
     LocatedNear & is located near \\
     RelatedTo & is related to \\
     InheritsFrom & inherits from \\
     LocationOfAction & is acted at the location of \\
     HasPainIntensity & causes pain intensity of \\
     \midrule
     AtLocation rev & is the position of \\
     CapableOf rev & is a skill of \\
     NotCapableOf rev & is not a skill of \\
     Causes rev & because \\
     CausesDesire rev & because \\
     CreatedBy rev & create \\
     DefinedAs rev & is known as \\
     DesireOf rev & is desired by \\
     Desires rev & is desired by \\
     NotDesires rev & is not desired by \\
     HasA rev & is possessed by \\
     HasFirstSubevent rev & is the beginning of \\
     HasLastSubevent rev & is the end of \\
     HasPrerequisite rev & is the prerequisite of \\
     HasProperty rev & is the property of \\
     InstanceOf rev & include \\
     IsA inversed & includes \\
     MadeOf rev & make up of \\
     MotivatedByGoal rev & motivate \\
     PartOf rev & include \\
     ReceivesAction rev & affect \\
     SymbolOf rev & can be represented by \\
     UsedFor rev & could make use of \\
     LocatedNear rev & is located near \\
     RelatedTo inversed & is related to \\
     InheritsFrom rev & hands down to \\
     LocationOfAction rev & is the location for acting \\
     HasPainIntensity rev & is the pain intensity caused by \\
    \bottomrule
    \end{tabular}
    \caption{Relation types and relation substitute templates from ConceptNet. \textit{rev} mean reverse relation.  }
    \label{append:2}
\end{table}

\subsection{Baselines}
\label{sec:appendix2}

\subsubsection{Commonsense Question Answering}


\noindent \textbf{Self-Talk \cite{shwartz2020unsupervised}}. Self-Talk inquires LMs for implicit background knowledge to solve multiple-choice commonsense tasks. The model uses pretrained LMs as knowledge sources. \\

\noindent \textbf{Dou \cite{dou2022zero}}. Dou extracts the related background knowledge embedded in pre-trained LMs by designing fill-in-the-blank prompts for commonsense question answering tasks. We compare with the syntactic-based rewriting method in which no supervision from curated-annotated training data is used.

\noindent \textbf{KTL \cite{banerjee2020self}}. Two ways are used to learn from knowledge triplets from knowledge graphs which can be used to perform zero-shot QA, namely knowledge representation learning (KRL) and span masked language modeling (SMLM). We compare with the KRL method. \\

\noindent \textbf{COMET-DynaGen \cite{bosselut2021dynamic}.} COMET-DG implements zero-shot commonsense QA by inference over dynamically generated commonsense knowledge graphs as related knowledge from COMET. 

\subsubsection{Inductive CSKG Completion}

\noindent \textbf{ConvE \cite{dettmers2018convolutional}.} ConvE stacks the node embedding and relation embedding and reshapes the resulting tensor into the same dimensionality as the node embeddings by a 2D convolution operation. \\

\noindent \textbf{RotatE \cite{sun2018rotate}.} RotatE utilizes the rotation operation and defines the distance function between entities and relation as $d_{r}(\mathbf{h}, \mathbf{t})=\|\mathbf{h} \circ \mathbf{r}-\mathbf{t}\|$. \\

\noindent \textbf{Malaviya \cite{malaviya2020commonsense}.} The method adopts a graph convolutional neural network GCN to learn graph structure information and a pre-trained LMs to represent contextual knowledge. \\

\noindent \textbf{InductivE \cite{wang2021inductive}.} The model directly computes entity embeddings from raw attributes and a GCN decoder with a novel densification process to enhance unseen entity representation with neighboring structural information. 



\end{document}